\documentclass{SCIS2025}

\usepackage{tikz}
\usepackage{multirow}
\usepackage{enumitem} 
\usepackage{tikz}
\usepackage{multirow}
\usepackage{enumitem} 
\usepackage{mdframed}
\usepackage{xcolor}
\usepackage{amsmath} 
\usepackage{graphicx}  
\usepackage{multirow}  
\usepackage{booktabs}  
\usepackage{colortbl}  
\usepackage{array}     
\usepackage{xcolor}    
\usepackage{mdframed}
\usepackage{xcolor}
\usepackage{graphicx}
\usepackage{caption} 
\usepackage{cuted}   
\usepackage{lipsum}  
\usepackage{multirow, makecell} 

\begin{document}

\ArticleType{RESEARCH PAPER}
\Year{2024}
\Month{}
\Vol{}
\No{}
\DOI{}
\ArtNo{}
\ReceiveDate{}
\ReviseDate{}
\AcceptDate{}
\OnlineDate{}

\title{Multimodal priors-augmented text-driven 3D human-object interaction generation}{Multimodal priors-augmented text-driven 3D human-object interaction generation}

\author[1]{Yin Wang}{}
\author[1]{Ziyao Zhang}{}
\author[1]{Zhiying Leng}{}
\author[1]{Haitian Liu}{}
\author[2]{Frederick W. B. Li}{}
\author[1]{\\Mu Li}{}
\author[1,3,]{Xiaohui Liang}{{liang\_xiaohui@buaa.edu.cn}}

\AuthorMark{Yin Wang}

\AuthorCitation{Yin Wang, et al}


\address[1]{State Key Laboratory of Virtual Reality Technology and Systems, Beihang University, Beijing, China}
\address[2]{Department of Computer Science, University of Durham, U.K}
\address[3]{Zhongguancun Laboratory, Beijing, China}

\abstract{We address the challenging task of text-driven 3D human-object interaction (HOI) motion generation. Existing methods primarily rely on a direct text-to-HOI mapping, which suffers from three key limitations due to the significant cross-modality gap: (Q1) sub-optimal human motion, (Q2) unnatural object motion, and (Q3) weak interaction between humans and objects. 
To address these challenges, we propose MP-HOI, a novel framework grounded in four core insights:
(1) Multimodal Data Priors: We leverage multimodal data (text, image, pose/object) from large multimodal models as priors to guide HOI generation, which tackles Q1 and Q2 in data modeling.
(2) Enhanced Object Representation: We improve existing object representations by incorporating geometric keypoints, contact features, and dynamic properties, enabling expressive object representations, which tackles Q2 in data representation.
(3) Multimodal-Aware Mixture-of-Experts (MoE) Model: We propose a modality-aware MoE model for effective multimodal feature fusion paradigm, which tackles Q1 and Q2 in feature fusion.
(4) Cascaded Diffusion with Interaction Supervision: We design a cascaded diffusion framework that progressively refines human-object interaction features under dedicated supervision, which tackles Q3 in interaction refinement.
Comprehensive experiments demonstrate that MP-HOI outperforms existing approaches in generating high-fidelity and fine-grained HOI motions.}


\maketitle


\section{Introduction}

Humans continuously interact with surrounding objects in daily life, e.g., moving monitors onto desks, pushing suitcases to desired locations, or washing apples and placing them on plates. Each task requires precise interaction between humans actions and objects movements. Exploring human-object interaction motion generation holds significant importance due to its broad downstream applications in character animation, VR/AR content creation, and robotics \cite{sui2025survey,fan20253d,zhu2023human,guo2015adaptive,wang2026dynamic,li2025fine}. As language serves as a natural interface for expressing interaction intentions, text-driven HOI motion generation has emerged as a promising research direction, aiming to generate 3D human-object interaction sequences guided by text prompts.

Existing work has explored text-driven HOI motion generation using diffusion models \cite{ho2020denoising} guided by text \cite{lv2024himo,zeng2025chainhoi}, object trajectories \cite{li2023object,li2024controllable}, contact maps \cite{peng2023hoi,song2024hoianimator}, and other cues \cite{wu2024thor,cha2024text2hoi,zhang2024force}.
Despite progress, current methods remain limited to coarse-grained interaction motions, falling short in three key aspects: 
\textit{(Q1) Sub-optimal Human Motion.} Existing methods, such as HIMO-Gen~\cite{lv2024himo}, typically take text and object geometry as input and generate human motion through simple condition concatenation. This insufficient modeling of conditions leads to sub-optimal human motion, resulting in low-quality sequences that are misaligned with the intended interactions.  
\textit{(Q2) Unnatural Object Motion.} Prior works~\cite{lv2024himo,li2023object} often adopt a simplified object representation, namely 3D translation and 6D rotation (only 9 dimensions), which disregards the geometric structure of objects. Consequently, the generated object motions often appear unnatural, exhibiting floating or sliding artifacts.  
\textit{(Q3) Weak Interaction Motion.} Methods such as CHOIS~\cite{li2024controllable} employ a one-step and direct text-to-HOI mapping. However, given the complexity of human-object interactions, this approach struggles to capture fine-grained HOI dynamics, leading to unrealistic contact, interpenetration, or implausible interactions in the generated motions.


To address these challenges, our core insight for effective modeling of human–object interaction can be structured around four core innovations. 
\textbf{(1) Multimodal Data Priors — Addressing \textit{Q1} and \textit{Q2} in data modeling.} We extract structured hierarchical priors from large multimodal models. Textual (1D): fine-grained descriptions derived from language parsing, Visual (2D): image-based motion references, and Spatial (3D): human atomic actions and object geometric structure. These multimodal priors provide rich semantic guidance for generating both human and object motion. 
\textbf{(2) Enhanced Object Representation — Addressing \textit{Q2} in data representation.} We augment object representations with additional 3D mesh keypoints, contact information, and translational/angular velocities. By enriching objects with detailed geometric and dynamic attributes, we enable more precise and stable object motion generation. 
\textbf{(3) Multimodal-Aware Mixture-of-Experts Model — Addressing \textit{Q1} and \textit{Q2} in feature fusion.} We propose a modality-aware MoE framework that selects specialized experts to enable effective multimodal interaction, which optimizes the multimodal feature fusion paradigm.  
\textbf{(4) Cascaded Diffusion with Interaction Supervision — Addressing \textit{Q3} in interaction optimization.} We design a cascaded diffusion framework: human diffusion, object diffusion, and human–object interaction diffusion. Coarse-grained human and object motions are generated first to guide fine-grained interaction motion generation. Additional supervision losses further refine the interaction features.

\begin{figure*}[t]
    \centering
    \includegraphics[width=\linewidth]{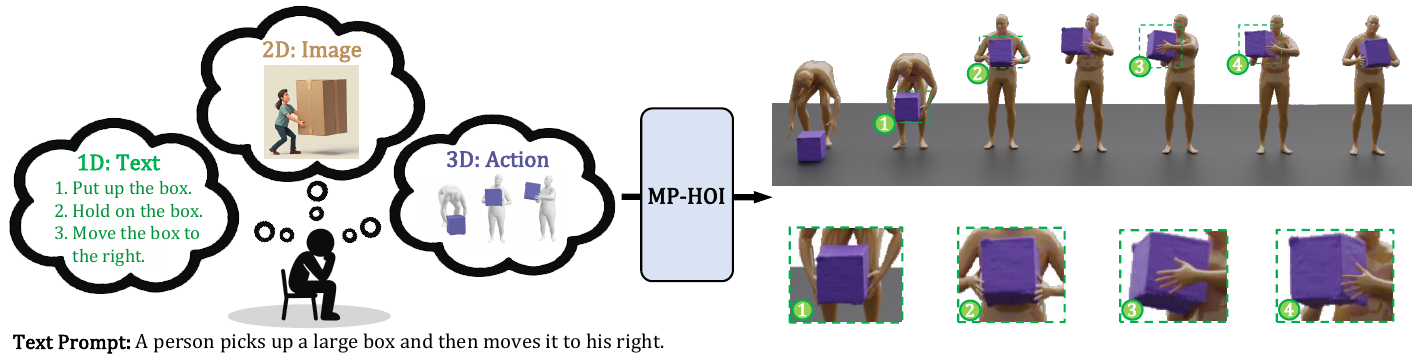}
    \caption{MP-HOI excels in generating fine-grained human-object interaction motions from multimodal data priors, achieving both high-quality human-object interactions and precise text-motion alignment.}
    \label{teaser}
\end{figure*}

We validate our approach through comprehensive experiments on two benchmark datasets: FullBodyManipulation (single-object interaction) \cite{li2023object} and HIMO (multi-objects interaction) \cite{lv2024himo}. Results show that our method outperforms existing techniques, achieving a new state-of-the-art in text-driven HOI motion generation. Our contributions are summarized as follows:

\begin{itemize}

\item We introduce the leverage of multimodal priors—1D textual, 2D visual, and 3D spatial—to guide the fine-grained human–object interaction motion generation.

\item We propose an enhanced object representation that incorporates geometric keypoints, contact features, and dynamic properties, enabling structurally rich and stable motion representation.

\item We present a modality-aware mixture-of-experts model that optimizes the fusion paradigm for multimodal features. 

\item We design a cascaded diffusion framework that progressively refines human–object interaction features under dedicated supervision, achieving state-of-the-art performance on existing benchmarks.

\end{itemize}

\section{Related Work}

\subsection{Text-Driven Human Motion Generation}
Three primary methodologies have emerged to tackle the challenge of text-driven human motion generation. (i) Latent Space Alignment \cite{ahuja2019language2pose,ghosh2021synthesis,tevet2022motionclip,petrovich2022temos,guo2022generating} aims to learn a unified latent space between text and motion embeddings. (ii) Conditional Autoregressive Models \cite{guo2022tm2t,zhang2023generating,zhong2023attt2m,lucas2022posegpt,jiang2023motiongpt,zhang2024motiongpt} generate motion tokens sequentially by leveraging previous tokens and text. In recent advancements \cite{pinyoanuntapong2025bamm,pinyoanuntapong2024mmm,guo2023momask} utilize masked motion modeling to generate more natural movements. 
(iii) Conditional Diffusion Models \cite{zhang2022motiondiffuse,tevet2023human,kim2023flame,chen2023executing,wang2023fg,zhang2023remodiffuse,zhang2025large,zhang2023finemogen,wang2025fg,wang2025most}, which learn probabilistic text-to-motion mappings within a conditional diffusion framework, have shown remarkable performance. 
While these advancements have propelled human motion generation forward, they predominantly center on individual motion generation, lacking the ability to generate interactive motions with external elements (e.g., objects).

\subsection{Text-Driven Human-Object Interaction Generation} 
The generation of human-object interactions has recently emerged as a promising research direction, attracting increasing attention from the community \cite{xu2024interdreamer,song2024hoianimator,peng2023hoi}. OMOMO \cite{li2023object} generates 3D human pose sequences based on the given motion of interacting 3D objects. GRAB \cite{taheri2020grab} predicts 3D hand grasping poses for specific 3D object shapes, performing various grasping manipulations. InterDiff \cite{xu2023interdiff} develops a diffusion-based generative model that predicts future human-object collaborative motion from their 3D interaction history. CG-HOI \cite{diller2024cg} proposes a method to generate realistic 3D human-object interactions from text descriptions and given static object geometry. IMoS \cite{ghosh2023imos} synthesizes full-body human and 3D object motions from textual inputs, but focuses exclusively on small-object grasping. HIMO \cite{lv2024himo} introduces a large-scale motion capture dataset of humans interacting with multiple objects and develops a baseline model. However, current methods can only achieve coarse-grained human-object interactions, which primarily manifest in three limitations: sub-optimal human motion, unnatural object motion, and weak interaction motion. Therefore, exploring fine-grained human-object interaction generation remains a critical yet challenging problem.

\subsection{Large Model-Assisted Motion Generation} 
In recent years, large language models (LLMs) such as BERT \cite{devlin2018bert}, GPT-4 \cite{achiam2023gpt}, and T5 \cite{raffel2020exploring} have demonstrated remarkable capabilities in language tasks. Some works have leveraged the strengths of LLMs to assist in motion generation tasks. For example, 
ActionGPT \cite{kalakonda2023action} utilizes GPT-3 to parse input text prompts into simple and long-form text prompts. SINC \cite{athanasiou2023sinc} also employs GPT-3.5 to establish relationships between motion and the human body. FineMoGen \cite{zhang2023finemogen} uses LLMs to adjust text prompts according to user input requirements for motion editing tasks. Fg-T2M++ \cite{wang2025fg} leverages GPT-4 to parse text prompts into detailed prompts for body part joints and analyzes the keyword properties in the text prompts to enable fine-grained motion generation.
Overall, current methods primarily utilize the text generation capability of LLMs to enhance the richness of textual prompts, thereby improving motion generation tasks.
However, they only exploit the single-modality (text) prior knowledge of LLMs, lacking consideration for leveraging multimodal large models (e.g., incorporating image) to effectively guide motion generation tasks.

\section{Preliminarily}
\textbf{Mixture-of-Experts} \cite{jacobs1991adaptive,eigen2013learning} assigns specific tasks to specialized experts, each adept at handling a particular aspect of the problem. This approach is well-suited to our multimodal priors-augmented HOI task, where the fusion of multimodal content and motion features presents a dynamic and complex challenge. 
Mixture-of-Experts involves a gating network to activate distinct subsets of expert networks for different inputs, which mainly consists of two key components. (1) MoE Layer:  A MoE layer contains $N$ experts (denoted as $e_i(\cdot), i=1,2,\ldots, N$). (2) Gating Network: A router $G$ routes the input token $\boldsymbol{x}$ to the most suitable top-$k$ experts. Formally, given the input token $\boldsymbol{x}$, the output token $\boldsymbol{y}$ of the MoE layer is the weighted sum of outputs from the $k$ activated experts:
\begin{equation}\label{eq:moe}
\boldsymbol{y}=\sum_{i\in\mathcal{T}}g_i(\boldsymbol{x})e_i(\boldsymbol{x}),
\end{equation}
where $g(\boldsymbol{x})=\sigma(\boldsymbol{W}\boldsymbol{x})$, where $\boldsymbol{W}$ is the gate parameter, $\sigma$ is the softmax function, and $\mathcal{T}$ represent the set of the top-$k$ indices.

Training MoE models directly often results in most tokens being assigned to a few experts, while others do not get  sufficient training. Therefore, a load balancing loss \cite{zoph2022st,lepikhin2020gshard} is applied to ensure a balanced distribution of input tokens across the experts:
\begin{equation}
\mathcal{L}_\text{b}=\sum_{i=1}^Nf_iP_i,
\end{equation}
where $f_i = \frac{1}{T_t} \sum_{t=1}^{T_t} \boldsymbol{1}(\text{Token } \boldsymbol{x}_t \text{ selects Expert } i)$, $f_i$ represents the fraction of tokens routed to expert $i$. $P_i$ is the fraction of the router probability assigned to expert $i$, defined as $P_i = \frac{1}{T_t} \sum_{t=1}^{T_t} \text{Softmax}\big(g(\boldsymbol{x}_t)\big)_i$, $T_t$ denotes the number of tokens, $N$ is the number of experts, $\boldsymbol{1}(*)$ denotes the indicator function.

\begin{figure*}[t]
    \centering
    \includegraphics[width=\linewidth]{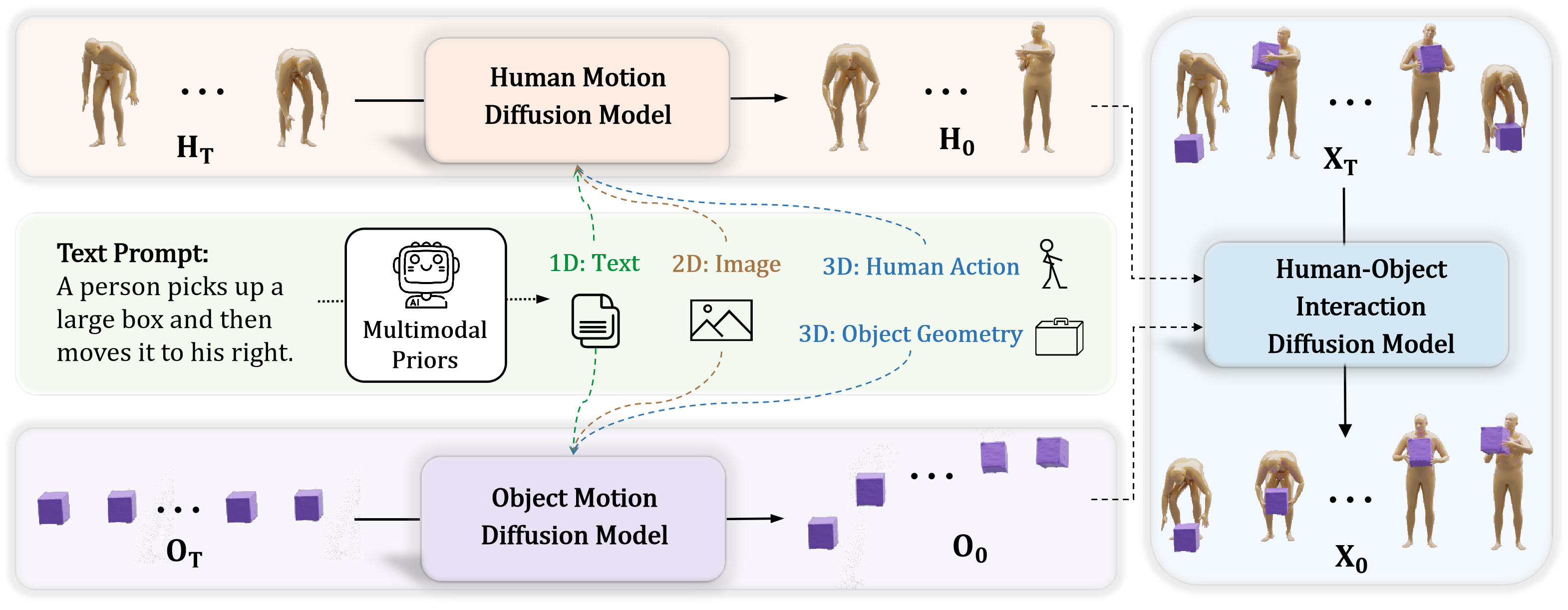}
    \caption{\textbf{Overview of MP-HOI.} Given a text prompt and multimodal priors, the reverse denoising process of the Human Motion Diffusion Model and Object Motion Diffusion Model starts from noisy motion data $H_T$ and $O_T$, generating clean human and object motion data ($H_0$ and $O_0$). Then, the Human-Object Interaction Diffusion Model takes the text prompt and the clean human and object motion data ($H_0$ and $O_0$) as inputs, and generates the final clean human-object interaction motion data $X_0$.}
    \label{pipeline}
\end{figure*}

\section{Methodology}

\subsection{Overview} 
We first formulate the text-driven HOI generation task. Given a text prompt $\mathbf{T}_p$ and object geometry $\mathbf{G}_o$, our goal is to generate a HOI motion sequence $\mathbf{M} \in \mathbb{R}^{S \times D}$, where $S$ indicates the length of the motion sequence and $D$ represents the dimension of HOI motion representation, which comprises a human motion $\mathbf{H}$ and an object motion $\mathbf{O}$. 
The text prompt $\mathbf{T}_p$ is represented as $\mathbf{T}_p \in \mathbb{R}^{N \times L}$, where $N$ denotes the number of words and $L$ is the dimension of the word vector. 
The object geometry $\mathbf{G}_o$ is represented as $\mathbf{G}_o \in \mathbb{R}^{K \times 3}$, where $K$ denotes the number of vertices on object mesh. 

As illustrated in Figure \ref{pipeline}, we introduce MP-HOI, a diffusion model-based framework for text-driven HOI generation. 
We start from a human motion random noisy and an object motion random noisy, represented respectively as $\mathbf{H}_{T} \in \mathbb{R}^{S \times D_h}$ and $\mathbf{O}_{T} \in \mathbb{R}^{S \times D_o}$, where $D_h$ denotes the dimension of human motion, $D_o$ represents the dimension of object motion.
MP-HOI utilizes a Human Motion Diffusion Model and an Object Motion Diffusion Model to denoise them over $t_h$ and $t_o$ steps, respectively, generating $\mathbf{H}_{0}$ and $\mathbf{O}_{0}$. 
These are then fed as conditional inputs into a Human-Object Interaction Diffusion Model to facilitate the denoising process of the HOI motion random noise, ultimately producing $\mathbf{X}_{0}$ over $t_{\text{hoi}}$ steps.

\subsection{Data Representation}

\textbf{Human Representation.} We denote the human motion as $\mathbf{H} \in \mathbb{R}^{S \times D_h}$. We adopt the SMPL-X \cite{pavlakos2019expressive} parametric model to represent the human motions. The representation $D_h$ consists of the global joint position $D_h^{j} \in \mathbb{R}^{52 \times 3}$, joint rotation $D_h^{r} \in \mathbb{R}^{52 \times 6}$ represented in the continuous 6D rotation format \cite{zhou2019continuity} and global translation $D_h^{t} \in \mathbb{R}^{3}$. 
Thus, the overall human motion representation is 471 dimensions. Notably, since the FullBodyManipulation  \cite{li2023object} dataset does not provide hand parameters, we omit the hand component of the SMPL-X \cite{pavlakos2019expressive} model during processing.

\textbf{Enhanced Object Representation.} We denote Object motion as $\mathbf{O} = \{O_j\}_{j=0}^{N_o} \in \mathbb{R}^{S \times D_o}$, where $N_o$ represents the number of the objects. Typically, each $D_o$ in $O_j$ includes relative rotation $D_o^{r} \in \mathbb{R}^{6}$ and global translation $D_o^{t} \in \mathbb{R}^{3}$. Therefore, the object motion representation contains only 9 dims in total, which is significantly fewer than the human motion representation. This dimension gap poses a substantial challenge for learning object motion. 
To address this limitation, we enhance the object representation with three additional informative features. First, we incorporate the object's translational velocity $D_o^{v_t} \in \mathbb{R}^{3}$ and angular velocity $D_o^{v_a} \in \mathbb{R}^{3}$, both derived from its translation and rotation. 
Second, we represent the object's geometric point cloud in a reduced form using 51 key points (50 sampled points plus 1 centroid). The global positions of these points $D_o^{p} \in \mathbb{R}^{51\times3}$ are computed via translation and rotation. 
Third, we include contact label information $D_o^{c} \in \mathbb{R}^{2}$ by calculating the distance between the object and the human's left and right hands. 
As a result, our enhanced object representation $D_o$ comprises a total of 170 dimensions per object.

\begin{figure*}[t]
    \centering
    \includegraphics[width=\linewidth]{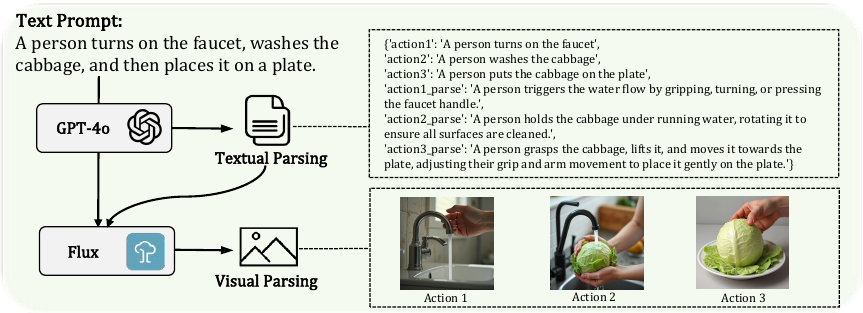}
    \caption{The pipeline for large models processing multimodal data (text and image).}
    \label{GPT}
\end{figure*}

\subsection{Multimodal Priors}
Existing methods predominantly adopt a direct text-to-HOI mapping approach for generation. However, due to the significant cross-modal gap between text and motion modalities, such methods often exhibit limitations in modeling fine-grained HOI interactions.
Recently, large-scale models across various modalities (e.g., GPT-4 \cite{achiam2023gpt}, DeepSeek \cite{liu2024deepseek}, DALL-E 3 \cite{betker2023improving}, Sora\cite{liu2024sora}) have advanced the field of multimodal learning through their powerful modeling capabilities. Our key insight is to leverage multimodal priors from diverse modalities (textual (1D), visual (2D), and spatial (3D)) to enhance the model's comprehension of interaction concepts, providing rich semantic guidance for generating human/object motion.

For the textual 1D prior, we leverage the strong priors of GPT-4o \cite{achiam2023gpt} to accurately capture the fine-grained relationship between natural language and human motion. Given a text prompt, we parse the sequence of actions being performed and provide detailed semantic explanations for each action. Specifically, for an input text such as ``A person turned on the faucet, washed the cabbage, and then places it on a plate," we first analyze the execution order of motions to obtain action-1, action-2, and action-3. Then, we provide fine-grained descriptions for action-i to support detailed understanding of interactive concepts at the textual level. The complete prompt is provided in the supplementary materials.

For the visual 2D prior, we leverage the powerful text-to-image capabilities of Flux \cite{flux2024} to provide fine-grained visual guidance for HOI. In the previous stage, we have already extracted a sequence of interaction actions—for example, action-1 is ``A person turned on the faucet.'' Each action-i is then used as a textual prompt, augmented with style keywords such as ``realism, photographic, detailed hand'' and fed into Flux to generate the corresponding HOI images. These visual cues, generated for each interaction step, further strengthen the understanding of human-object interactions at the visual level. A complete example is provided in Figure \ref{GPT}. 

For the spatial 3D prior, we process the human and object components separately. For the object, we regard its geometric structure as a rich source of spatial information and use its point cloud data $G_o \in \mathbb{R}^{1024\times3}$ as the 3D prior. For the human, we observe that HOI types are not infinitely diverse but instead fall into a limited set. Through all datasets \cite{li2023object,lv2024himo} analysis, we identified approximately 55 common HOI actions, such as pick up, place, eat, wash and so on. For each category, we select the most representative frame that captures the essence of the interaction as an atomic motion, and pair it with a corresponding textual annotation. 
To associate a given action-i with the most relevant atomic motion, we perform text-based retrieval based on semantic similarity. Specifically, we use the CLIP \cite{radford2021learning} to extract text features for both the action-i text and all annotated atomic motion texts, and compute cosine similarity to find the closest match. By integrating the object's spatial geometry with the human's representative atomic motion, we further reinforce the understanding of HOI at the spatial level.

In summary, given a text prompt, we extract fine-grained action-parsed textual descriptions $C_t$, visual images $C_v$ generated based on the interaction order text, atomic motions $C_a$, and object point clouds $G_o$. For feature extraction, we employ the CLIP to encode the text $C_t$ and the images $C_v$, yielding feature representations $C_t^f \in \mathbb{R}^{N_a \times D_t}$ and $C_v^f \in \mathbb{R}^{N_a \times D_v}$, respectively, where $N_a$ denotes the number of text or image. The atomic motions $C_a$ are encoded using a MLP to obtain $C_a^f \in \mathbb{R}^{N_a \times D_a}$, while the object point clouds $G_o$ are processed with a PointNet \cite{qi2017pointnet} architecture to obtain $C_p^f \in \mathbb{R}^{N_b \times D_p}$, where $N_b$ denotes the number of object geometry.

\begin{figure*}[t]
    \centering
    \includegraphics[width=\linewidth]{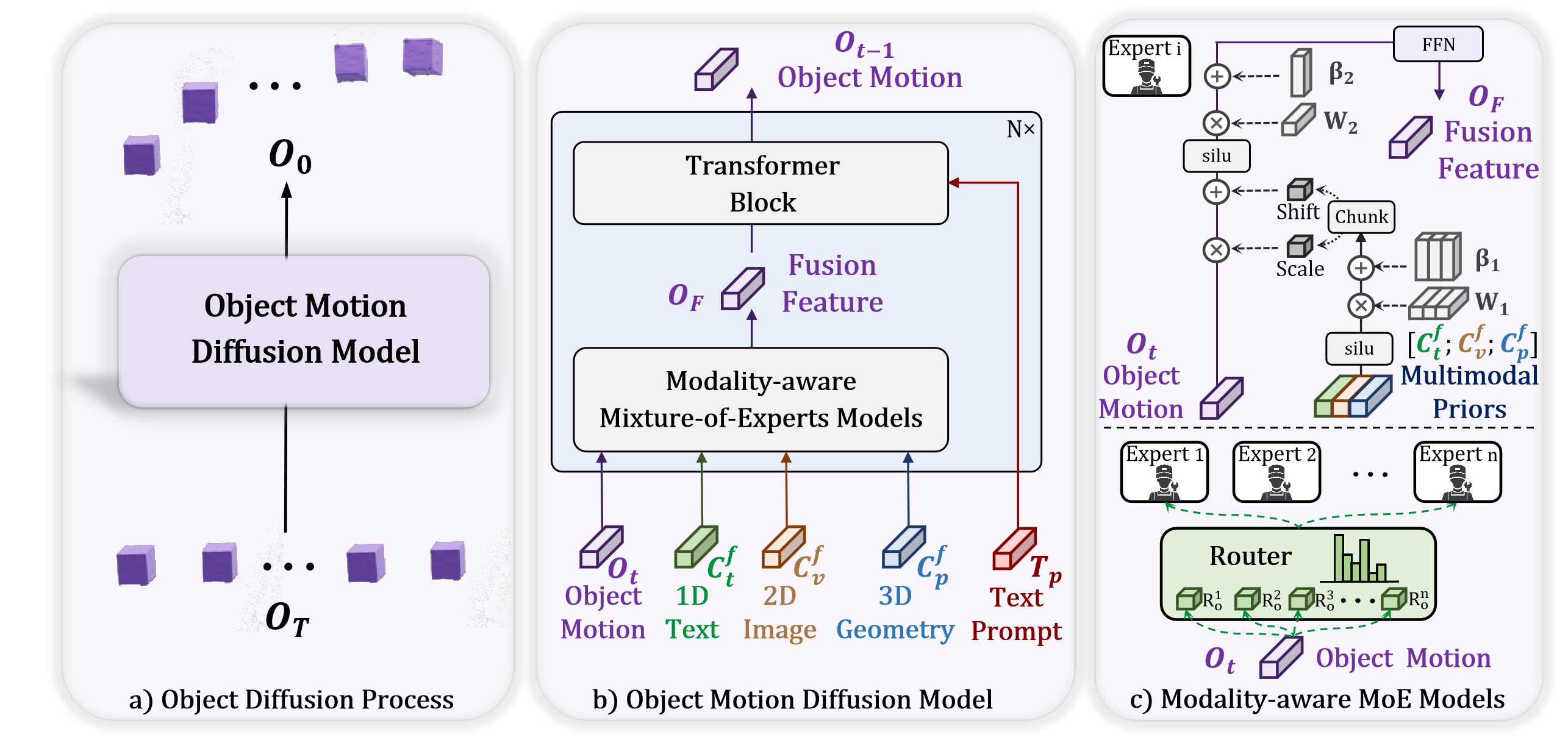}
    \caption{\textbf{Illustration of the overall object motion diffusion pipeline.} (a) Object diffusion process. (b) Object motion diffusion model. (c) Architecture of Modality-aware MoE Models. Notably, the Human Motion Diffusion Model adopts this identical architecture, with the object geometry feature $C_p^f$ replaced by the atomic motion feature $C_a^f$.}
    \label{object_diffusion}
\end{figure*}

\subsection{Human/Object Motion Diffusion Process} 
Existing methods typically adopt a one-step generation approach (e.g., only the Human-Object Interaction Diffusion in Figure \ref{pipeline}) to generate human-object interaction motions. However, due to the inherent complexity of HOI dynamics, such direct text-to-HOI mapping often yields suboptimal results, including imprecise human motion, unnatural object trajectories, and coarse-grained interaction patterns.
To address this issue, we propose a multi-step generation paradigm to enhance interaction quality. As illustrated in Figure \ref{pipeline}, our framework first employs separate Human Motion Diffusion and Object Motion Diffusion model to generate preliminary human and object motions, conditioned on text prompts and multimodal priors. These intermediate motions serve as bridging representations, mitigating the feature gap between textual descriptions and HOI sequences. Subsequently, the preliminary human and object motions are integrated into the Human-Object Interaction Diffusion Model as reference features to guide the HOI generation process, ultimately yielding refined HOI motion sequences.

\paragraph{Human/Object Motion Diffusion Model} 
Figure~\ref{object_diffusion} presents the complete object motion diffusion pipeline. It is worth noting that the human motion diffusion pipeline follows a similar process, with the only difference being the replacement of the object geometry feature $C_p^f$ with the atomic motion feature $C_a^f$. Therefore, we take the object motion diffusion model as an example for detailed explanation.

The goal of the Object Motion Diffusion Model is to generate a plausible object motion sequence based on text prompts and multimodal priors. We take the denoising process at time step $t$ as an example, as illustrated in Figure \ref{object_diffusion}(b). Specifically, the object motion features $O_t$ and multimodal prior features ($C_t^f, C_v^f, C_p^f$) are fed into a Modality-aware Mixture-of-Experts Model which serves as an information fusion module. This module injects the multimodal knowledge into the object motion features, producing a fused representation. The fused object motion features are then combined with the text prompt features $T_p$ and passed through multiple transformer blocks, each consisting of a self-attention layer, a cross-attention layer, and a feedforward layer. This process ultimately outputs the refined object motion features $O_{t-1}$ at time step $t–1$.

\paragraph{Modality-aware Mixture-of-Experts Models} 

To generate fine-grained human-object interaction motion using data priors that provide rich semantic guidance, the core challenge lies in effectively fusing multi-modal condition features with HOI motion features. On a deeper level, the fundamental heterogeneity across different modalities (e.g., textual, visual, and spatial) introduces additional significant challenges for feature fusion. 
General fusion methods, such as simple concatenation or attention-based fusion, are typically static and inflexible. They may prioritize certain modalities while neglecting the information gains from others, failing to preserve fine-grained semantic features within some modalities features that are crucial for capturing the nuances of human-object interaction modeling.

Mixture of Experts ~\cite{fan2022m3vit,chen2023adamv} offers a promising alternative by dynamically routing inputs to specialized experts, each handling distinct modality patterns.
Thus, we argue that MoEs provides a highly effective and flexible solution to this challenge.
However, previous MoE models typically adopt simple FFNs as expert layers~\cite{fedus2022switch,du2022glam}. Such architectures are insufficient for effectively integrating motion and multi-modal condition features, as they fail to account for how each modality influences motion representation in a fine-grained manner. To address this limitation, we propose the Modality-aware Mixture-of-Experts model, as shown in Figure \ref{object_diffusion} (c).


Specifically, the MoEs includes a routing function that directs the input features to the appropriate expert models, and several experts specializing in different tasks. Taking the object motion feature $O_t$ as an example, the routing function contains a routing parameter matrix $R_o$ that is used to assign tasks:
\begin{equation}
\mathbf{l}_{o} = \tau_{t} (\mathrm{W}_{o}\mathbf{O}_t) R_{o},
\end{equation}
where $\tau_{t}$ is the trainable temperature hyper-parameter, $\mathrm{W}_{o}$ is the trainable matrix, $\mathbf{l}_{o} \in \mathbb{R}^{N_t \times N_e}$ represents the logits for selecting experts in object motion. Here, $N_t$ denotes the number of tokens input to the MoE, and $N_e$ denotes the number of expert models. By considering which expert model best matches the input tokens, the Router dynamically identifies and selects the most suitable expert.

As for expert architecture, drawing inspiration from FiLM \cite{perez2018film}, we design the condition modulation module to adaptively model the influence of multi-modal conditions on motion features: 
\begin{equation}
[\mathbf{C}_f^1, \mathbf{C}_f^2] = \mathrm{W}_{1} \phi( [\mathbf{C}_t^f;\mathbf{C}_v^f;\mathbf{C}_a^f]) + \beta_{1},
\end{equation}
where $\mathrm{W}_{1}$ and $\beta_{1}$ are the trainable matrices, $[\cdot ; \cdot]$ indicates a concatenation of input tensors, $\phi$ denotes the activation function, $\mathbf{C}_{f}^{1} \in \mathbb{R}^{1 \times D_{f}}$ and $\mathbf{C}_{f}^{2} \in \mathbb{R}^{1 \times D_{f}}$ represent the scale and shift parameters, respectively, which are then adaptively fused with the motion features:
\begin{equation}
\mathbf{O}_F = \text{FFN} [ \mathrm{W}_{2} \phi [(1+\mathbf{C}_{f}^{1})\mathbf{O}_t + \mathbf{C}_{f}^{2}] + \beta_{2}],
\end{equation}
where $\mathrm{W}_{2}$ and $\beta_{2}$ are the trainable matrices, $\mathbf{O}_F$ is then reshaped to $\mathbb{R}^{T \times D_{o}}$ as the final fusion output of the MoE Layer.

\paragraph{Human/Object Motion Diffusion Training Objective}
To supervise human/object motion generation process, we minimize the L2 loss between predicted and ground truth motions, denoted as 
\begin{equation}
\mathcal{L}_{h/o}^{l2}=\mathbb{E} [\parallel \mathbf{x}_0 - \epsilon_\theta(\mathbf{x}_t,t,\mathbf{T}_p,C_t^f, C_v^f, C_{a/p}^f) \parallel_2^2] .
\label{L2loss}
\end{equation}
where $\epsilon_\theta(\mathbf{x}_t,t,\mathbf{T}_p,C_t^f, C_v^f, C_{a/p}^f)$ denotes the model prediction, $\mathcal{L}_{h}^{l2}$ and $\mathcal{L}_{o}^{l2}$ denote the human motion diffusion loss and the object motion diffusion loss, respectively. In addition, it is necessary to consider the load balancing loss of the MoE, which serves to optimize the expert assignment mechanism. Therefore, the total training loss in this section includes the $\mathcal{L}_{1}$ loss and load balancing loss: $\mathcal{L}_{h/o} = \mathcal{L}_{h/o}^{l2} + \lambda_{b} \mathcal{L}_{b}$, where $\lambda_{b}$ is the trade-off hyperparameter.

\subsection{Human-Object Interaction Diffusion Process}

\begin{figure*}[t]
    \centering
    \includegraphics[width=\linewidth]{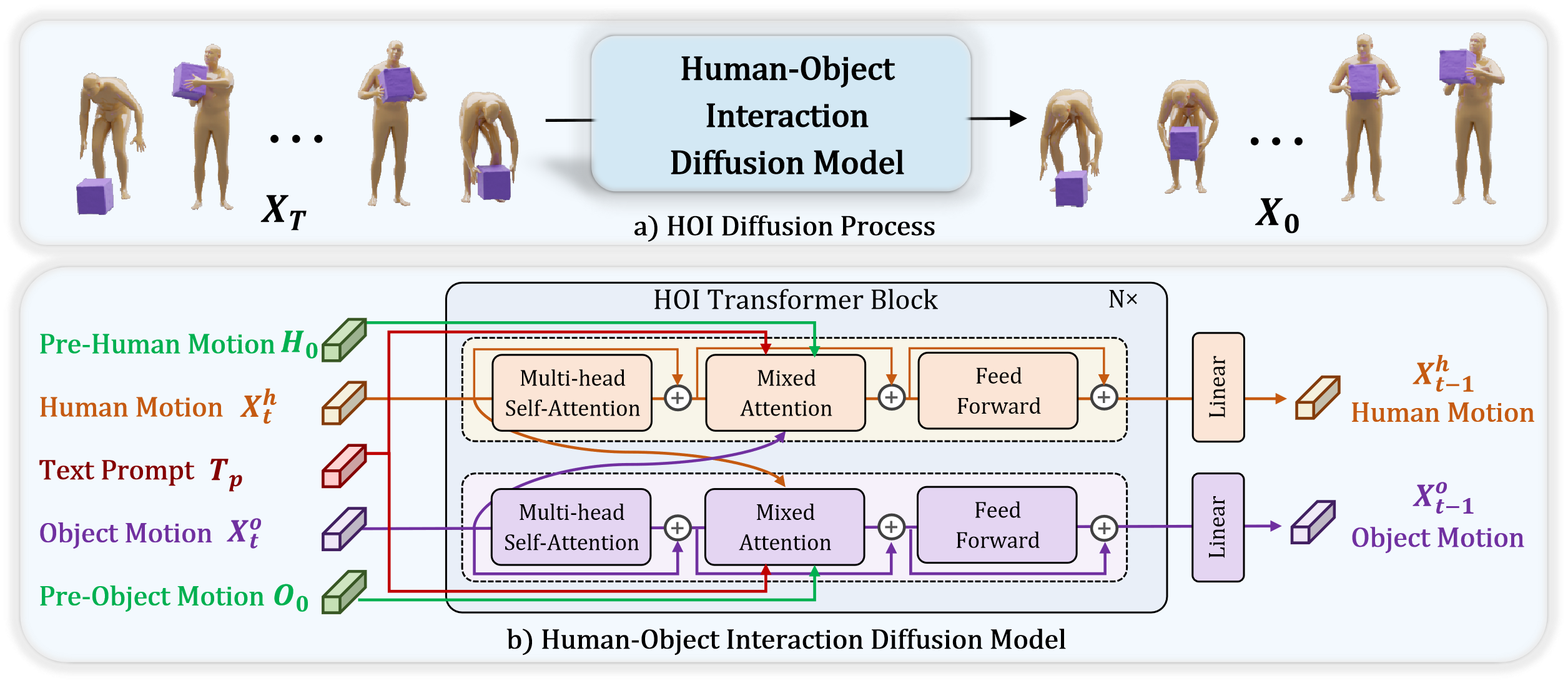}
    \caption{\textbf{Illustration of the overall human-object interaction motion diffusion pipeline.} (a) Human-object interaction motion diffusion process. (b) Architecture of human-object interaction diffusion model. Pre-Human Motion and Pre-Object Motion represent the human motion and object motion generated in the human/object motion diffusion process, respectively.}
    \label{interaction_diffusion}
\end{figure*}

\paragraph{Human-Object Interaction Diffusion Model}
The Human-Object Interaction Diffusion Model generates a fine-grained human-object interaction motion sequence based on the text prompt and the preliminary human and object motions produced in the previous stage, as illustrated in Figure \ref{interaction_diffusion}. We take the denoising process at time step $t$ and human motion $\mathbf{X}_t^h$ as an example. Specifically, the human motion features $\mathbf{X}_t^h$ are passed through multiple HOI transformer blocks, each consisting of a self-attention layer, a mixed-attention layer, and a feedforward layer. The mixed-attention layer facilitates information exchange between the human motion $\mathbf{X}_t^h$ and various contextual features ($\mathbf{H}_0, \mathbf{T}_{p}, \mathbf{X}_t^o$). Formally, the Query, Key, and Value matrices in this layer are computed as follows:
\begin{equation}
\mathbf{Q}=\mathrm{Q}_{m}^{h} \mathbf{X}_t^h, \quad
\mathbf{K} = [\mathrm{K}_{h} \mathbf{H}_0; \mathrm{K}_{p} \mathbf{T}_{p}; \mathrm{K}_{m}^{o} \mathbf{X}_t^o], \quad
\mathbf{V} = [\mathrm{V}_{h} \mathbf{H}_0; \mathrm{V}_{p} \mathbf{T}_{p}; \mathrm{V}_{m}^{o} \mathbf{X}_t^o], 
\end{equation}
where $\mathrm{Q}_{m}^{h}$, $\mathrm{K}_{h}$, $\mathrm{K}_{p}$, $\mathrm{K}_{m}^{o}$, $\mathrm{V}_{h} $, $\mathrm{V}_{p}$ and $\mathrm{V}_{m}^{o}$ are trainable matrices. 
Then, the global templates $\mathbf{G_g}$ in the attention are computed to yield the output $\mathbf{Y}$:
\begin{equation}
\mathbf{G_g} = \operatorname{softmax} (\mathbf{K})\mathbf{V}, \quad
\mathbf{Y} = \operatorname{softmax} (\mathbf{Q})\mathbf{G_g}.
\end{equation}

By passing through multiple HOI transformer blocks, the human motion representation is progressively refined to produce the optimized motion feature $\mathbf{X}_{t-1}^h$. Similarly, the object motion $\mathbf{X}_t^o$ undergoes a parallel process, resulting in the optimized object motion feature $\mathbf{X}_{t-1}^o$.

\paragraph{Human-Object Interaction Diffusion Training Objective}
To supervise the human-object interaction motion generation process, we employ a comprehensive set of loss functions to ensure the plausibility of the generated motion. First, we apply an L2 loss $\mathcal{L}_{l2}$, similar to Equation \ref{L2loss}, to encourage accurate reconstruction of HOI motion. The HOI motion representation consists of both human and object motion components. Notably, the human motion representation includes global joint positions, joint rotations, and global translation, all of which are effectively constrained by the L2 loss to ensure high-quality human motion. More importantly, the object motion representation is enriched with additional features such as relative rotation, global translation, angular velocity, translational velocity, the global positions of keypoints, and contact information. Consequently, even this simple L2 loss implicitly supervises these diverse and informative object features.
Second, we introduce a velocity-level constraint $\mathcal{L}_{vel}$ by computing the velocities of the human body and both hands, ensuring that the generated motion exhibits realistic dynamics. Third, to maintain reasonable spatial relationships among objects, we incorporate a distance-based loss $\mathcal{L}_{dis}$ that penalizes physically implausible inter-object distances. Fourth, to prevent unnatural penetration between the human and the object, we apply an interaction loss $\mathcal{L}_{inter}$, which computes the distance between the human's left/right hands and the object's centroid, thereby guiding appropriate proximity in human-object interactions. In summary, we formulate the final optimization objective at this stage as:
\begin{equation}
\mathcal{L}_{\text{hoi}}= \lambda_{l2} \mathcal{L}_{l2} + \lambda_{vel} \mathcal{L}_{vel} + \lambda_{dis} \mathcal{L}_{dis} + \lambda_{inter} \mathcal{L}_{inter}
\end{equation}
where $\lambda_{l2}, \lambda_{vel}, \lambda_{dis}, \lambda_{inter}$ are hyperparameters. Since our MP-HOI is trained in an end-to-end manner, the overall training objective is defined as the sum of the losses from the preceding stages, formulated as: $\mathcal{L} = \mathcal{L}_{h} + \mathcal{L}_{o} + \mathcal{L}_{\text{hoi}}$.

\section{Experiments}

\begin{table*}[t]
\centering
\resizebox{\linewidth}{!}
{ 
\begin{tabular}{cllcccccc}
\toprule[1.25pt]
\multirow{2}{*}{Dataset} & \multirow{2}{*}{Methods} &\multirow{2}{*}{Publication} &\multicolumn{3}{c}{{Motion Quality Evaluation}} &\multicolumn{1}{c}{{Diversity Evaluation}} \\
\cmidrule(rl){4-6} \cmidrule(rl){7-7} 
&&&R-TOP 3 $\uparrow$&FID $\downarrow$ &MM Dist $\downarrow$&Diversity $\uparrow$ \\ \midrule
\multirow{5}{*}{\parbox{4cm}{\centering FullBodyManipulation\\ (1 Object)}}
&
OMOMO~\cite{li2023object} 
&TOG 2023
& $0.773^{\pm.009}$
& $1.276^{\pm.016}$
& $2.468^{\pm.034}$
& $9.719^{\pm.087}$
\\&
MotionDiffuse~\cite{zhang2022motiondiffuse} 
&TPAMI 2024
& $0.830^{\pm.003}$
& $0.892^{\pm.019}$
& $2.220^{\pm.021}$
& $\cellcolor{blue!10}10.02^{\pm.066}$
\\
&
CHOIS ~\cite{li2024controllable} 
&ECCV 2024
& $0.791^{\pm.005}$
& $\cellcolor{blue!10}0.823^{\pm.012}$
& $\cellcolor{blue!10}2.177^{\pm.011}$
& $9.998^{\pm.037}$
\\
&
HIMO-Gen~\cite{lv2024himo} 
&ECCV 2024
& $\cellcolor{blue!10}0.851^{\pm.008}$
& $0.924^{\pm.026}$
& $2.346^{\pm.035}$
& $9.877^{\pm.089}$
\\
\cmidrule(l){2-7} 
&Ours (MP-HOI) & -& $\cellcolor{red!10}\bf{0.872^{\pm.005}}$ & $\cellcolor{red!10}\bf{0.703^{\pm.042}}$& $\cellcolor{red!10}\bf{1.948^{\pm.016}}$ & $\cellcolor{red!10}\bf{10.38^{\pm.059}}$ \\

\midrule
\multirow{8}{*}{\parbox{4cm}{\centering HIMO\\ (2 Objects)}}&
IMoS~\cite{ghosh2023imos}
& CGF 2023
& $0.501^{\pm.012}$ 
& $7.589^{\pm.112}$ 
& $8.740^{\pm.031}$ 
& $7.003^{\pm.320}$ 
\\
&
MDM~\cite{tevet2023human} 
& ICLR 2023
& $0.605^{\pm.009}$ 
& $6.845^{\pm.331}$ 
& $8.018^{\pm.050}$ 
& $11.38^{\pm.234}$ 
\\
&
OMOMO~\cite{li2023object} 
& TOG 2023
& $0.592^{\pm.012}$ 
& $6.132^{\pm.271}$ 
& $7.921^{\pm.065}$ 
& $\cellcolor{blue!10}12.73^{\pm.196}$ 
\\
&
PriorMDM~\cite{shafir2023human} 
& ICLR 2024
& $0.589^{\pm.003}$ 
& $7.851^{\pm.251}$ 
& $7.250^{\pm.006}$ 
& $12.57^{\pm.146}$ 
\\
&
MotionDiffuse~\cite{zhang2022motiondiffuse}
& TPAMI 2024
& $0.576^{\pm.009}$ 
& $4.364^{\pm.039}$ 
& $5.190^{\pm.039}$ 
& $10.79^{\pm.106}$ 
\\
&
CHOIS ~\cite{li2024controllable} 
& ECCV 2024
& $0.567^{\pm.041}$ 
& $3.996^{\pm.587}$ 
& $5.986^{\pm.693}$ 
& $12.44^{\pm.514}$ 
\\
&
HIMO-Gen~\cite{lv2024himo} 
& ECCV 2024
& $\cellcolor{blue!10}0.636^{\pm.003}$ 
& $\cellcolor{blue!10}1.481^{\pm.042}$ 
& $\cellcolor{blue!10}3.649^{\pm.010}$ 
& $11.66^{\pm.204}$ 
\\
\cmidrule(l){2-7} 
&Ours (MP-HOI) & -& $\cellcolor{red!10}\bf{0.842^{\pm.007}}$ & $\cellcolor{red!10}\bf{1.070^{\pm.021}}$& $\cellcolor{red!10}\bf{2.968^{\pm.029}}$ & $\cellcolor{red!10}\bf{12.83^{\pm.079}}$ \\

\midrule
\multirow{8}{*}{\parbox{4cm}{\centering HIMO\\ (3 Objects)}}&
IMoS~\cite{ghosh2023imos}
& CGF 2023
& $0.466^{\pm.101}$ 
& $4.990^{\pm.177}$ 
& $7.770^{\pm.058}$ 
& $9.231^{\pm.113}$ 

\\&
MDM~\cite{tevet2023human} 
& ICLR 2023
& $0.502^{\pm.013}$ 
& $4.571^{\pm.110}$ 
& $6.314^{\pm.026}$ 
& $8.895^{\pm.285}$ 

\\
&
OMOMO~\cite{li2023object} 
& TOG 2023
& $0.553^{\pm.037}$ 
& $4.561^{\pm.039}$ 
& $5.463^{\pm.049}$ 
& $9.169^{\pm.073}$ 
\\
&
PriorMDM~\cite{shafir2023human} 
& ICLR 2024
& $0.513^{\pm.025}$ 
& $4.821^{\pm.203}$ 
& $5.890^{\pm.023}$ 
& $\cellcolor{blue!10}9.340^{\pm.023}$ 

\\
&
MotionDiffuse~\cite{zhang2022motiondiffuse} 
& TPAMI 2024
& $0.515^{\pm.013}$ 
& $4.719^{\pm.059}$ 
& $5.673^{\pm.046}$ 
& $8.993^{\pm.097}$ 

\\
&
CHOIS ~\cite{li2024controllable} 
& ECCV 2024
& $\cellcolor{blue!10}0.602^{\pm.007}$ 
& $\cellcolor{blue!10}3.653^{\pm.046}$ 
& $\cellcolor{blue!10}4.763^{\pm.046}$ 
& $9.135^{\pm.091}$ 
\\
&
HIMO-Gen~\cite{lv2024himo} 
& ECCV 2024
& $0.535^{\pm.018}$ 
& $4.771^{\pm.110}$ 
& $5.086^{\pm.041}$ 
& $8.946^{\pm.137}$ 

\\
\cmidrule(l){2-7} 
&Ours (MP-HOI) & - & $\cellcolor{red!10}\bf{0.729^{\pm.009}}$ & $\cellcolor{red!10}\bf{1.621^{\pm.030}}$& $\cellcolor{red!10}\bf{3.392^{\pm.019}}$ & $\cellcolor{red!10}\bf{10.02^{\pm.103}}$  \\

\bottomrule[1.25pt]
\end{tabular}
}
\caption{Comparisons to current state-of-the-art methods on the FullBodyManipulation \cite{li2023object} and HIMO \cite{lv2024himo} test set. ``$\uparrow$'' denotes that higher is better. ``$\downarrow$'' denotes that lower is better. 
We repeat all the evaluations 20 times and report the average with a 95\% confidence interval. 
We report the best and the second-best results in \textcolor{red}{red} cells and \textcolor{blue}{blue} cells.}
\label{main_compare}
\end{table*}

\subsection{Datasets, Metrics and Implementation Details}

\textbf{Datasets.} 
\textbf{FullBodyManipulation} \cite{li2023object} contains 10 hours of motion data involving human interactions with a \textbf{single object}, comprising a total of 4,838 HOI sequences. Each HOI sequence is accompanied by a textual description that guided the volunteers during the motion recording. A total of 17 participants were involved in the data collection process. They interacted with each object according to the given textual instructions. The dataset includes 15 commonly used objects in daily tasks, such as clothes stand, suitcase, table, trashcan, monitor, and others.

\textbf{HIMO} \cite{lv2024himo} includes 9.44 hours of motion data depicting human interactions with \textbf{two or three objects}, comprising 3,376 HOI sequences. Each sequence is paired with a textual description. A total of 34 participants contributed to the data collection. The dataset covers 53 everyday household objects, such as plate, laptop, bottle, apple, bowl, and more. It also encompasses many interaction types, including: Put A (and B) into C, Wash A (and B) under faucet, Use A and B, Place A on B, among others.

\textbf{Metrics.}
To evaluate HOI motion generation, we first employ general metrics to assess the quality of the generation, such as \textbf{R-TOP}, \textbf{FID}, \textbf{MM-Dist}, and \textbf{Diversity}. \textbf{R-TOP} reflects the semantic consistency between generated HOIs and the given textual prompts.
\textbf{FID} measures the similarity between the feature distributions extracted from the generated motions and the ground truth motions. \textbf{MM-Dist} computes the average Euclidean distance between the feature of generated motions and the text prompt feature. \textbf{Diversity} evaluates the dissimilarity among all generated motions across all descriptions. 
Furthermore, we follow \cite{li2023object,li2024controllable} evaluation metrics to assess the quality of human-object interactions, including \textbf{Interaction Distance}, \textbf{Contact Percentage}, \textbf{Precision}, \textbf{Recall} and \textbf{F1 score}. We first compute the distance between hand positions and the object centroid point. The interaction distance $D_I$ is defined as the difference between the distances of the generated motions and those of the ground truth motions. Additionally, a contact threshold is empirically set to determine contact labels for each frame. We then count true positives, false positives, and false negatives to compute precision $C_{prec}$, recall $C_{rec}$, and F1 score $C_{F1}$. Moreover, Contact Percentage $C_{\%}$ reflects frame-level contact inference accuracy and is defined as the proportion of frames where contact is detected.

\textbf{Implementation Details.}
Regarding the multimodal priors, we utilize GPT-4o \cite{achiam2023gpt} to process text data and Flux \cite{flux2024} to generate image data. The hyperparameter $N_a$ is set to 2 on the FullBodyManipulation \cite{li2023object} dataset and 3 on the HIMO \cite{lv2024himo} dataset. $N_b$ is set to 1 on the FullBodyManipulation (1 object), 2 on HIMO (2 objects), and 3 on HIMO (3 objects).
Regarding the motion diffusion model, we employ a 4-layer transformer in human motion diffusion model, a 4-layer transformer in object motion diffusion model and a 8-layer transformer in HOI motion diffusion model. As for the MoE models, the number of experts is set to 16, with top-k set to 2. The dimension $R_0$ is set to 256. As for the text encoder, a frozen text encoder from CLIP ViT-B/32 is utilized, complemented by two additional transformer encoder layers. Regarding some hyperparameters, $D_a$ is set to 512, $D_p$ is set to 256, $\lambda_b$ is set to 10, $\lambda_{l2}$ is set to 1, $\lambda_{vel}$ is set to 2, $\lambda_{dis}$ is set to 0.3, $\lambda_{inter}$ is set to 0.01, the guidance scale is set to 2.0. In terms of the diffusion model, the variances $\beta_t$ are predefined to linearly spread from 0.0001 to 0.02, and the total number of noising steps is set at T = 1000. We use the Adam optimizer to train the model with an initial learning rate of 0.0001, gradually decreasing to 0.00001 through a cosine learning rate scheduler. The training process is conducted on 4 NVIDIA GeForce RTX 3090, with a batch size of 16 on a single GPU.

\begin{table*}[t]
\centering
\resizebox{\linewidth}{!}
{ 
\begin{tabular}{clcccccccc}
\toprule[1.25pt]
\multirow{2}{*}{Dataset} &\multirow{2}{*}{Methods} &\multirow{2}{*}{Publication}&\multicolumn{5}{c}{{Human-Object Interaction Evaluation}}\\
\cmidrule(rl){4-8} 
  && &$C_{prec}$ $\uparrow$ &$C_{rec}$ $\uparrow$ &$C_{F1}$ $\uparrow$ &$C_{\%}$ $\rightarrow$ &$D_I$ $\rightarrow$   \\ \midrule
\multirow{5}{*}{\parbox{4cm}{\centering FullBodyManipulation\\ (1 Object)}}
&
GT
& -
& -
& -
& -
& $0.445^{\pm 0.000}$
& $0.501^{\pm 0.000}$
\\
&
MotionDiffuse~\cite{zhang2022motiondiffuse} 
& TPAMI 2024
& $0.358^{\pm 0.015}$
& $0.272^{\pm 0.013}$
& $0.278^{\pm 0.013}$
& \cellcolor{blue!10}$0.321^{\pm 0.017}$
& $0.615^{\pm 0.019}$
\\
&
CHOIS ~\cite{li2024controllable} 
& ECCV 2024
& \cellcolor{blue!10}$0.372^{\pm 0.009}$
& \cellcolor{blue!10}$0.305^{\pm 0.009}$
& \cellcolor{blue!10}$0.303^{\pm 0.008}$
& $0.318^{\pm 0.019}$
& $0.637^{\pm 0.019}$
\\
&
HIMO-Gen~\cite{lv2024himo} 
& ECCV 2024
& $0.364^{\pm 0.008}$
& $0.286^{\pm 0.008}$
& $0.289^{\pm 0.009}$
& $0.297^{\pm 0.011}$
& \cellcolor{blue!10}$0.613^{\pm 0.013}$
\\
\cmidrule(l){2-8} 
&Ours (MP-HOI) &-
& \cellcolor{red!10}$\mathbf{0.396}^{\pm 0.007}$
& \cellcolor{red!10}$\mathbf{0.343}^{\pm 0.006}$
& \cellcolor{red!10}$\mathbf{0.342}^{\pm 0.007}$
& \cellcolor{red!10}$\mathbf{0.371}^{\pm 0.005}$
& \cellcolor{red!10}$\mathbf{0.589}^{\pm 0.007}$
\\

\midrule
\multirow{5}{*}{\parbox{4cm}{\centering HIMO\\ (2 Objects)}}
&
GT
& -
& -
& -
& -
& $0.833^{\pm 0.000}$
& $0.215^{\pm 0.000}$
\\
&
MotionDiffuse~\cite{zhang2022motiondiffuse} 
& TPAMI 2024
& \cellcolor{blue!10}$0.845^{\pm 0.013}$
& $0.764^{\pm 0.013}$
& $0.770^{\pm 0.012}$
& $0.747^{\pm 0.017}$
& $0.315^{\pm 0.017}$
\\
&
CHOIS ~\cite{li2024controllable} 
& ECCV 2024
& $0.829^{\pm 0.011}$
& $0.789^{\pm 0.012}$
& $0.791^{\pm 0.011}$
& \cellcolor{blue!10}$0.756^{\pm 0.010}$
& \cellcolor{blue!10}$0.294^{\pm 0.009}$
\\
&
HIMO-Gen~\cite{lv2024himo} 
& ECCV 2024
& $0.844^{\pm 0.007}$
& \cellcolor{blue!10}$0.804^{\pm 0.008}$
& \cellcolor{blue!10}$0.802^{\pm 0.007}$
& $0.707^{\pm 0.010}$
& $0.302^{\pm 0.009}$
\\
\cmidrule(l){2-8} 
&Ours (MP-HOI) &-
& \cellcolor{red!10}$\mathbf{0.863}^{\pm 0.010}$
& \cellcolor{red!10}$\mathbf{0.837}^{\pm 0.011}$
& \cellcolor{red!10}$\mathbf{0.835}^{\pm 0.009}$
& \cellcolor{red!10}$\mathbf{0.815}^{\pm 0.007}$
& \cellcolor{red!10}$\mathbf{0.255}^{\pm 0.008}$
\\

\midrule
\multirow{5}{*}{\parbox{4cm}{\centering HIMO\\ (3 Objects)}}
&
GT
& -
& -
& -
& -
& $0.843^{\pm 0.000}$
& $0.222^{\pm 0.000}$
\\
&
MotionDiffuse~\cite{zhang2022motiondiffuse} 
& TPAMI 2024
& $0.841^{\pm 0.015}$
& $0.795^{\pm 0.015}$
& $0.794^{\pm 0.016}$
& \cellcolor{blue!10}$0.784^{\pm 0.011}$
& \cellcolor{blue!10}$0.303^{\pm 0.014}$
\\
&
CHOIS ~\cite{li2024controllable} 
& ECCV 2024
& \cellcolor{blue!10}$0.850^{\pm 0.010}$
& \cellcolor{blue!10}$0.803^{\pm 0.011}$
& \cellcolor{blue!10}$0.804^{\pm 0.011}$
& $0.771^{\pm 0.015}$
& $0.332^{\pm 0.012}$
\\
&
HIMO-Gen~\cite{lv2024himo} 
& ECCV 2024
& $0.844^{\pm 0.010}$
& $0.772^{\pm 0.009}$
& $0.779^{\pm 0.010}$
& $0.758^{\pm 0.011}$
& $0.315^{\pm 0.014}$
\\
\cmidrule(l){2-8} 
&Ours (MP-HOI) &-
& \cellcolor{red!10}$\mathbf{0.859}^{\pm 0.009}$
& \cellcolor{red!10}$\mathbf{0.824}^{\pm 0.009}$
& \cellcolor{red!10}$\mathbf{0.825}^{\pm 0.008}$
& \cellcolor{red!10}$\mathbf{0.815}^{\pm 0.008}$
& \cellcolor{red!10}$\mathbf{0.274}^{\pm 0.007}$
\\

\bottomrule[1.25pt]
\end{tabular}
}
\caption{Human-object interaction evaluation on the FullBodyManipulation \cite{li2023object} and HIMO \cite{lv2024himo} test sets. `$\rightarrow$' means the closer to GT the better. }
\label{hoi_compare}
\end{table*}

\subsection{Evaluation of Human-Object Interaction Generation}

\paragraph{General Evaluation} 
The results in Table \ref{main_compare} compare MP-HOI against state-of-the-art (SOTA) methods including IMoS~\cite{ghosh2023imos}, MDM~\cite{tevet2023human}, OMOMO~\cite{li2023object}, PriorMDM~\cite{shafir2023human}, MotionDiffuse~\cite{zhang2022motiondiffuse}, CHOIS~\cite{li2024controllable}, and HIMO-Gen~\cite{lv2024himo} on the FullBodyManipulation \cite{li2023object} and HIMO \cite{lv2024himo} datasets.
In terms of motion quality evaluation, compared to other methods, MP-HOI achieves significantly higher scores in R-TOP3, FID, and MM-Dist. These results highlight our method's proficiency in generating high-quality HOI motion sequences that seamlessly align with the textual prompts. Notably, compared with the HIMO-Gen~\cite{lv2024himo}, our method reduces the FID by 23.92\% on the FullBodyManipulation single-object interaction dataset, by 27.75\% on HIMO 2-object dataset, and by an impressive 66.02\% on the 3-object dataset. 
We further analyze the underlying reasons. Specifically, MotionDiffuse~\cite{zhang2022motiondiffuse} and MDM~\cite{tevet2023human} rely primarily on single-modality inputs and do not incorporate object-related information. Without explicit modeling of object conditions, their generated motions often lack accurate and physically consistent human-object interactions. In contrast, IMoS~\cite{ghosh2023imos}, CHOIS~\cite{li2024controllable}, and HIMO-Gen~\cite{lv2024himo} consider object conditions, but they fail to accurately model interactions. Because these methods lack prior knowledge about human-object interactions, they cannot achieve a fine-grained understanding of interaction concepts. Furthermore, their object motion representations rely on a direct 9-dimensional encoding, which makes it difficult for the models to fully interpret the embedded information, resulting in lower motion quality. This limitation restricts their ability to model complex human-object interactions, which explains the significant performance gap compared to MP-HOI.


\begin{figure*}[t]
    \centering
    \includegraphics[width=\linewidth]{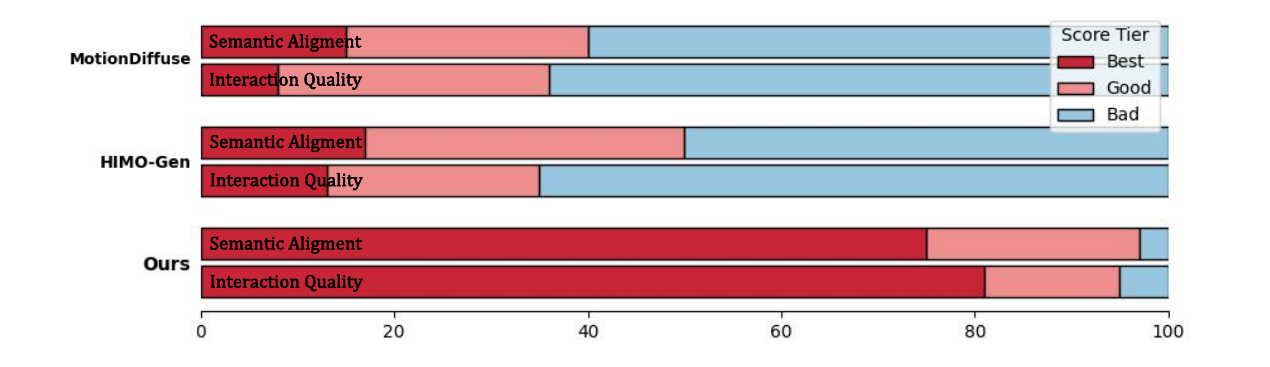}
    \vspace{-0.45cm}
    \caption{User study results. The color bars indicate the percentage distribution of scores for each evaluation criterion.}
    \label{userstudy}
\end{figure*}
\begin{table*}[t]
\centering
\resizebox{\linewidth}{!}
{ 
\begin{tabular}{lccccccccccc}
\toprule[1.25pt]
\multirow{2}{*}{Methods}  &\multicolumn{3}{c}{{Motion Quality Evaluation}} &\multicolumn{1}{c}{{Diversity Evaluation}} &\multicolumn{5}{c}{{Human-Object Interaction Evaluation}}\\
\cmidrule(rl){2-4} \cmidrule(rl){5-5}  \cmidrule(rl){6-10}
&R-TOP 3 $\uparrow$&FID $\downarrow$ &MM Dist $\downarrow$&Diversity $\uparrow$&$C_{prec}$ $\uparrow$ &$C_{rec}$ $\uparrow$ &$C_{F1}$ $\uparrow$ &$C_{\%}$ $\rightarrow$ &$D_I$ $\rightarrow$ \\ \midrule

w/o Multimodal Priors (1D, 2D, 3D)
& $0.761^{\pm.010}$
& $1.190^{\pm.022}$
& $2.321^{\pm.009}$
& $9.399^{\pm.019}$
& $0.369^{\pm.009}$
& $0.300^{\pm.008}$
& $0.304^{\pm.009}$
& $0.319^{\pm.031}$
& $0.602^{\pm.027}$
\\

\quad\quad w/o Multimodal Priors (1D)
& $0.784^{\pm.015}$
& $0.995^{\pm.019}$
& $2.150^{\pm.017}$
& $9.431^{\pm.045}$
& $0.386^{\pm.024}$
& $0.319^{\pm.026}$
& $0.321^{\pm.024}$
& $0.351^{\pm.035}$
& $0.582^{\pm.010}$
\\
\quad\quad w/o Multimodal Priors (2D)
& $0.772^{\pm.009}$
& $1.054^{\pm.011}$
& $2.207^{\pm.026}$
& $9.543^{\pm.027}$
& $0.381^{\pm.019}$
& $0.310^{\pm.020}$
& $0.309^{\pm.022}$
& $0.334^{\pm.025}$
& $0.597^{\pm.034}$
\\
\quad\quad w/o Multimodal Priors (3D)
& $0.769^{\pm.011}$
& $1.086^{\pm.026}$
& $2.226^{\pm.010}$
& $9.561^{\pm.029}$
& $0.373^{\pm.013}$
& $0.303^{\pm.012}$
& $0.301^{\pm.012}$
& $0.323^{\pm.019}$
& $0.599^{\pm.021}$
\\
GPT-4o + Flux
& $0.783^{\pm.008}$
& $1.516^{\pm.030}$
& $2.442^{\pm.019}$
& $9.576^{\pm.059}$
& $0.345^{\pm.018}$
& $0.312^{\pm.009}$
& $0.316^{\pm.010}$
& $0.311^{\pm.022}$
& $0.553^{\pm.044}$
\\
GPT-3.5 + Stable Diffusion
& $0.762^{\pm.010}$
& $1.811^{\pm.022}$
& $2.679^{\pm.046}$
& $9.421^{\pm.089}$
& $0.312^{\pm.025}$
& $0.298^{\pm.012}$
& $0.301^{\pm.016}$
& $0.307^{\pm.031}$
& $0.539^{\pm.024}$
\\
\midrule
w/o Enhanced Object Representation
& $0.759^{\pm.005}$
& $1.346^{\pm.011}$
& $2.475^{\pm.020}$
& $9.416^{\pm.069}$
& $0.359^{\pm.007}$
& $0.289^{\pm.008}$
& $0.290^{\pm.008}$
& $0.315^{\pm.022}$
& $0.615^{\pm.017}$
\\

w/o Cascaded Diffusion Framework
& $0.755^{\pm.005}$
& $1.368^{\pm.016}$
& $2.599^{\pm.030}$
& $9.711^{\pm.105}$
& $0.352^{\pm.012}$
& $0.310^{\pm.012}$
& $0.315^{\pm.012}$
& $0.309^{\pm.017}$
& $0.586^{\pm.030}$
\\

w/o Modality-aware MoE Models
& $0.836^{\pm.007}$
& $0.931^{\pm.020}$
& $2.080^{\pm.041}$
& $9.404^{\pm.085}$
& $0.379^{\pm.006}$
& $0.327^{\pm.007}$
& $0.329^{\pm.008}$
& $0.330^{\pm.009}$
& $0.597^{\pm.016}$
\\

w/o HOI Supervision Loss
& $0.841^{\pm.009}$
& $0.890^{\pm.019}$
& $2.174^{\pm.009}$
& $9.835^{\pm.052}$
& $0.364^{\pm.009}$
& $0.317^{\pm.009}$
& $0.319^{\pm.008}$
& $0.325^{\pm.011}$
& $0.569^{\pm.010}$
\\
\midrule
Ours (MP-HOI) & $\cellcolor{red!10}\bf{0.872^{\pm.005}}$ & $\cellcolor{red!10}\bf{0.703^{\pm.042}}$& $\cellcolor{red!10}\bf{1.948^{\pm.016}}$ & $\cellcolor{red!10}\bf{10.38^{\pm.059}}$
& $\cellcolor{red!10}\bf{0.396}^{\pm.007}$ & $\cellcolor{red!10}\bf{0.343}^{\pm.006}$& $\cellcolor{red!10}\bf{0.342}^{\pm.007}$ & $\cellcolor{red!10}\bf{0.371}^{\pm.005}$ &$\cellcolor{red!10}\bf{0.589}^{\pm.007}$\\

\bottomrule[1.25pt]
\end{tabular}
}
\caption{Ablation study results on the FullBodyManipulation \cite{li2023object} test set.}
\label{Ablation}
\end{table*}

\paragraph{Human-Object Interaction Evaluation}
We perform a Human-Object Interaction Evaluation to assess the interaction performance based on five HOI interaction metrics.
As shown in Table~\ref{hoi_compare}, compared to SOTA methods, MP-HOI achieves significant improvements in Precision $C_{prec}$, Recall $C_{rec}$, and F1 score $C_{F1}$. These results reveal that our method has better hand-object contact accuracy and enhanced physical reasonableness. For example, compared with CHOIS ~\cite{li2024controllable} in the 2-object setting of the HIMO dataset, $C_{prec}$ increases by 2.25\%, $C_{rec}$ increases by 4.10\%, and $C_{F1}$ increases by 4.12\%. 
In addition, our method also performs well on the Contact Percentage ($C_{\%}$) metric. For example, it improves by 15.27\% compared to HIMO-Gen~\cite{lv2024himo}, indicating that the percentage of contact frames in the MP-HOI generated human-object interaction motions is closest to that of the ground truth motions.
Finally, regarding the interaction distance ($D_{I}$) metric, our method achieves a 19.04\% reduction compared to MotionDiffuse~\cite{zhang2022motiondiffuse}. 
Therefore, the human-object distances in our generated motions are the most reasonable, effectively reducing interpenetration and other unrealistic artifacts.

\paragraph{User Study Evaluation}
We conduct a user study, in which we compare our method with HIMO-Gen~\cite{lv2024himo} and MotionDiffuse~\cite{zhang2022motiondiffuse}. This user study engaged 30 participants to evaluate 10 motion sequences generated by each method. The designed questionnaire consisted of two questions: (1) ``Which method generates motion that best aligns with the textual prompt?'' and (2) ``Which method best captures the fine-grained details of human-object interactions?'' 
Participants rated the methods on a 1-to-3 scale (indicating bad, good, and best). As depicted in Figure \ref{userstudy}, MP-HOI significantly outperforms the competing methods in both semantic alignment and interaction quality. These results demonstrate that MP-HOI not only produces motions that closely follow textual prompts but also excels in modeling human-object interactions, validating the effectiveness and superiority of our approach.


\begin{figure*}[t]
    \centering
    \includegraphics[width=\linewidth]{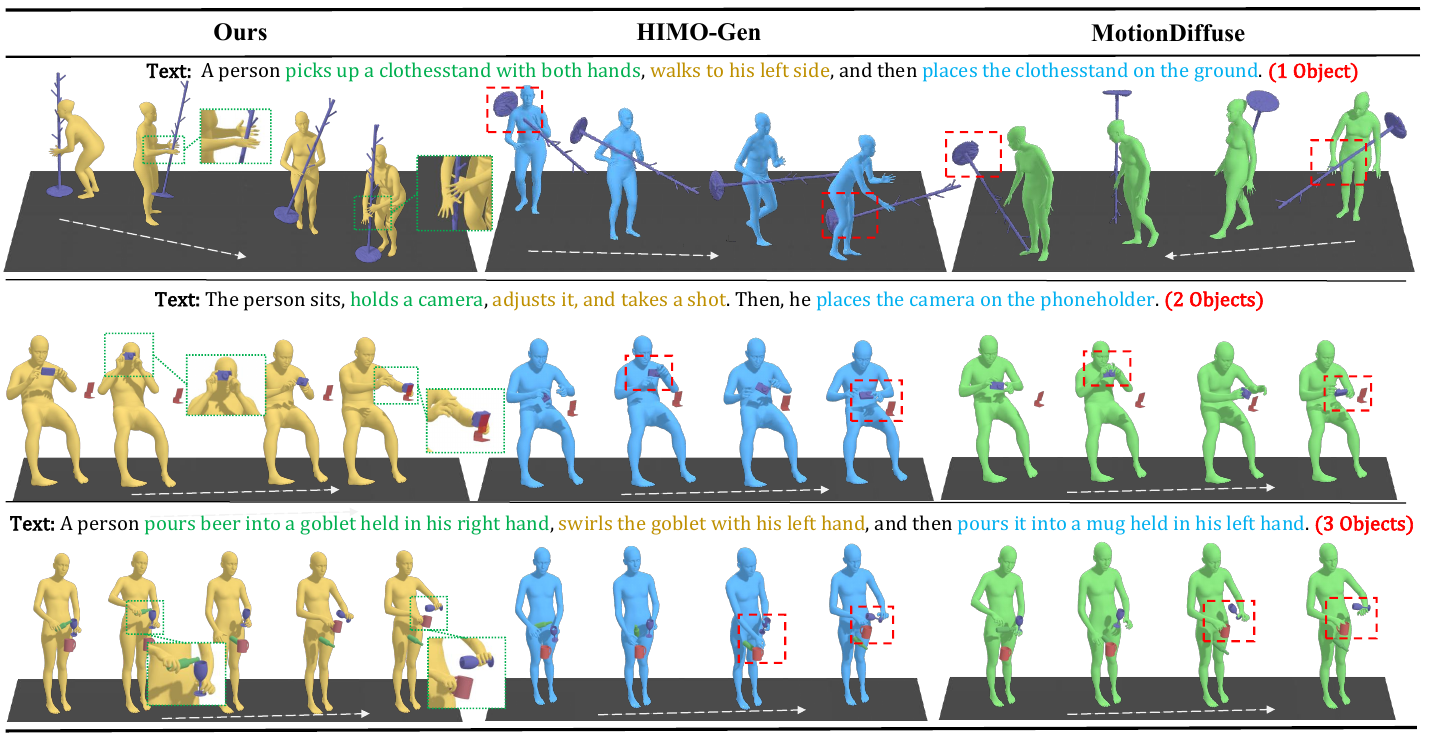}
    \caption{\textbf{Visual results compared with existing methods}. The arrow represents the time axes. The green box zooms in on the detailed interactions demonstrated by our approach. The red boxes highlight the errors in other methods.}
    \label{compare}
\end{figure*}

\paragraph{Qualitative Analysis}
Figure \ref{compare} qualitatively compares PerMoGen against HIMO-Gen \cite{lv2024himo} and MotionDiffuse \cite{zhang2022motiondiffuse}. 
In the single-object scenario, HIMO-Gen \cite{lv2024himo} exhibits significant human-object penetration and generates unrealistic object motion. Similarly, MotionDiffuse \cite{zhang2022motiondiffuse} fails to accurately follow the text prompts, often producing motion in the opposite direction of what is described.
In the two-object scenario, HIMO-Gen \cite{lv2024himo} generates low-quality motion where the camera is not properly held in the human's hand as instructed, while MotionDiffuse \cite{zhang2022motiondiffuse} fails to place the camera back onto the phoneholder as required by the text.
In the three-object scenario, both HIMO-Gen \cite{lv2024himo} and MotionDiffuse \cite{zhang2022motiondiffuse} demonstrate weak and imprecise interactions, struggling to engage with each object in a smooth and meaningful way.
In contrast, MP-HOI consistently produces motion that closely aligns with the textual prompts across all scenarios. It demonstrates fine-grained modeling of human-object interactions, highlighting the superiority of our approach.

\paragraph{Quantitative Ablation Study}
We conduct a comprehensive set of ablation studies to evaluate the contribution of each key component in MP-HOI, as summarized in Table~\ref{Ablation}. 
\textbf{First}, we examine the impact of multimodal priors. Removing the multimodal priors leads to a noticeable decline of 69.27\% in FID and 9.45\% in motion diversity. This result highlights the importance of incorporating multimodal cues, as they not only guide the generation of high-quality interactive motions but also significantly enhance motion diversity.
We further investigated the impact of each modality on the final performance and found that different modalities affect motion quality to varying degrees. For example, under the MM-Dist metric, removing the 1D prior leads to a 10.37\% drop in performance, removing the 2D prior results in a 13.29\% drop, and removing the 3D prior causes a 17.27\% drop. These results suggest that, since the task involves 3D motion generation, knowledge that is closer to the 3D domain more effectively guides the generation process.
Additionally, we examined the effect of different large model combinations on motion generation performance. We randomly sampled 100 examples and processed them using various combinations of large models. The results show that the combination of GPT-4o and Flux leads to more significant improvements in metrics, primarily because it provides more detailed and comprehensive text analysis and image-based prompts.
\textbf{Second}, when replacing our enhanced object encoding with the original 9-dimensional format (translation + rotation), we observe a substantial decline in HOI motion quality. Specifically, the FID increases by 91.46\%, and R-TOP 3 drops by 12.95\%, demonstrating that our enriched object representation is critical for generating high-quality HOI motions.
\textbf{Third}, eliminating the cascaded diffusion framework reduces MP-HOI to a single-step generation paradigm, similar to other baseline methods. This results in coarse and less realistic HOI motions, as reflected in the degradation of both motion quality and interaction accuracy. For instance, MM-Dist decreases by 33.41\%, and $C_{prec}$ drops by 11.11\%.
\textbf{Fourth}, when the modality-aware MoE module is removed, the model struggles to effectively fuse multimodal priors with HOI motion features. Thus, it fails to fully exploit the rich information across modalities, leading to noticeable drops in both motion quality and diversity, for example, a 32.43\% degradation in FID and a 9.40\% reduction in diversity.
\textbf{Finally}, removing the HOI supervision constraints results in poor human-object interaction quality. In particular, the $D_I$ metric experiences a significant performance drop (3.39\%), indicating less plausible spatial relationships and a higher frequency of interpenetration artifacts.
\begin{figure*}[t]
    \centering
    \includegraphics[width=\linewidth]{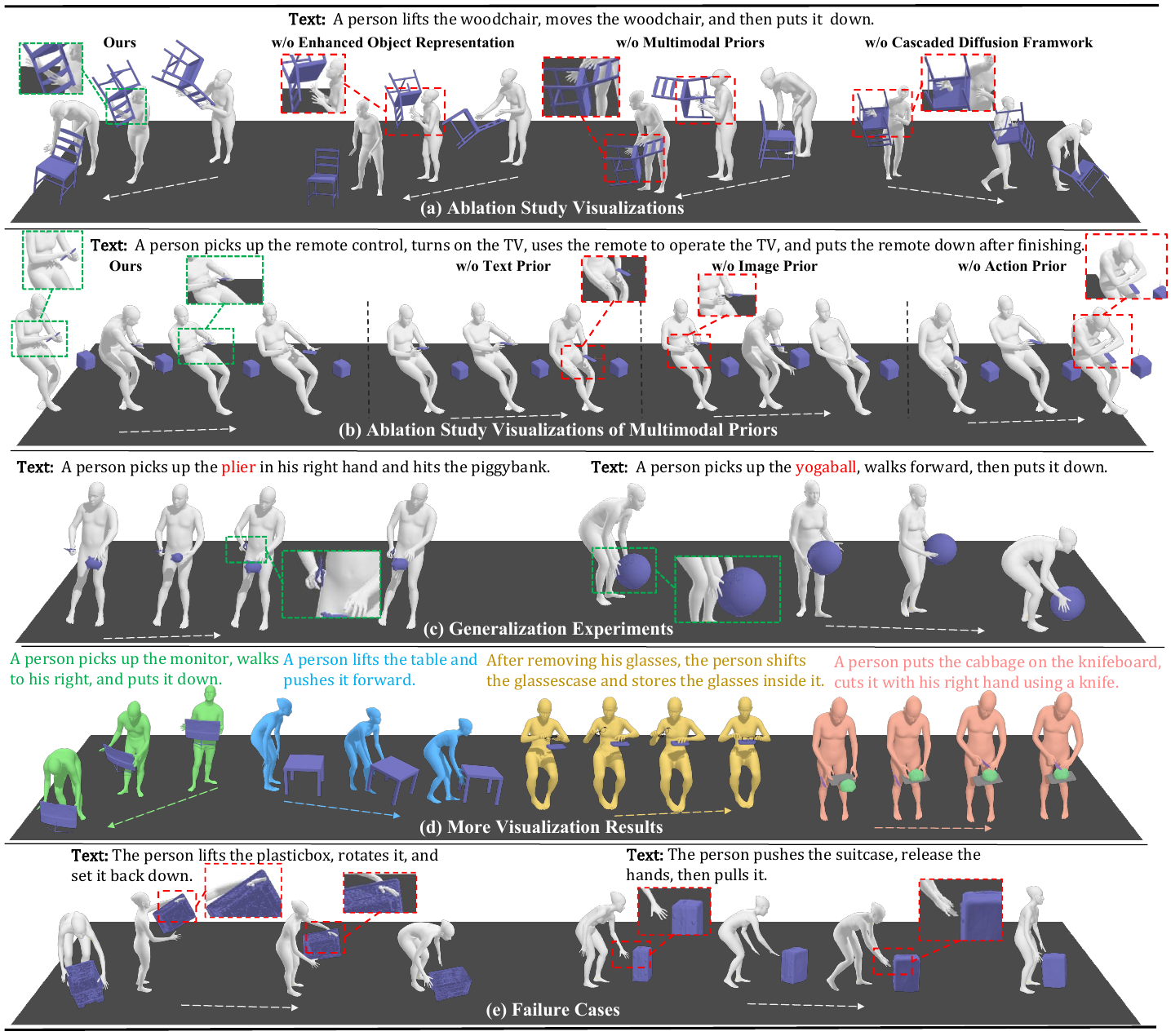}
    \caption{\textbf{Additional visualization results}. (a) Ablation study visualizations. (b) Ablation Study Visualizations of Multimodal Priors. (c) Generalization experiments. (d) More visualization results. (e) Failure cases. }
    \label{morevis}
\end{figure*}

\paragraph{Qualitative Ablation Study}
To thoroughly evaluate the individual contribution of each module, we employed motion visualization for an in-depth comparative analysis and examined the impact of removing key components, as shown in Figure~\ref{morevis} (a). When the enhanced object representation was removed, the quality of object motion significantly deteriorated, leading to sliding and floating artifacts of the chair. Without the multimodal priors, the human demonstrated poor interaction with the object, erroneously grasping unreasonable parts of the chair. In the absence of the cascaded diffusion framework, both the human and the motion quality degraded drastically, resulting in severe human-object penetration. In contrast, the complete MP-HOI model successfully performed the motion dictated by the text prompt and exhibited fine-grained human-object interaction, thereby validating the effectiveness of our approach.

Moreover, we provide ablation and visualization experiments for each modality, as shown in Figure \ref{morevis} (b). From these results, we observe that when the textual modality is removed, the generated motions may fail to fully reflect the multiple steps described in the input text. When the visual modality is removed, the quality of hand-object contact information is weakened. When the motion modality is removed, the generated motions tend to have lower overall motion quality. These observations demonstrate the specific contribution of each modality in supporting fine-grained HOI generation.

\paragraph{Generalization}
To evaluate the generalization ability of MP-HOI to unseen objects, we qualitatively assess the effectiveness of our method. We feed their geometries as input conditions into our model and generate corresponding human-object motions, as illustrated in Figure~\ref{morevis} (c).
In the first example, the human correctly grasps the end of the pliers and uses the tool appropriately. 
In the second example, the human interacts with a reasonable region of the yoga ball and successfully completes a sequence of actions, including picking it up, walking, and putting it down.
These examples demonstrate that MP-HOI can still generate high-quality human-object interaction motions even when interacting with previously unseen object geometries, such as a plier or a yoga ball, which are not present in the dataset. This highlights the effectiveness of our method and its strong generalization capability to unseen objects.

\paragraph{Additional Visualization Results}
Figure \ref{morevis} (d) presents four additional visualizations showcasing human-object interactions.
From the first and second examples on the left side, it is evident that MP-HOI can stably grasp appropriate parts of the monitor and the edge of the table, thereby accurately performing the actions described in the text. In the third example, MP-HOI effectively models the interaction between the human and two objects, successfully completing the task of picking up the glasses and placing them back into the glasses case. Even in the fourth example, which involves interaction with three objects, the human is able to reasonably manage the relationships among multiple objects and sequentially carry out the instructions from the text, demonstrating strong multi-object interaction modeling capabilities.
Further comparisons and visualization examples are provided in the supplementary video.

\paragraph{Failure Cases}
While MP-HOI effectively generates human-object interaction motion based on textual prompts, it encounters two failure cases shown in Figure \ref{morevis} (e). The primary failures occur on the FullBodyManipulation \cite{li2023object} dataset, which does not provide hand pose parameters. As a result, the generated motions on this dataset sometimes cannot accurately model finger positions, leading to potential hand-object penetrations. Additionally, during close-range interactions with objects, such as pushing a suitcase, the lack of detailed hand parameters may result in overly large contact distances between the hand and the object.

\section{Limitation and Future Work}
Here, we discuss the limitations of MP-HOI and suggest directions for future work. First, although the generated motion sequences generally align well with textual prompts, some artifacts, such as foot sliding, may occasionally occur. These issues could be alleviated in the future by incorporating additional physical constraints and enhancing physical simulations.
Second, our current framework is limited to interactions with rigid objects. However, in the real world, humans often interact with deformable or fluid-like substances, such as gases or liquids. Exploring human interactions with such dynamic materials presents an interesting and promising direction for future research.

\section{Conclusion}
In this paper, we address the challenging task of text-driven human-object interaction generation by introducing MP-HOI, a novel framework built upon four key insights:  
(1) We leverage multimodal priors, including text, images, and pose/object information, to effectively guide the HOI generation process;  
(2) We enhance object representations by incorporating geometric keypoints, contact features, and dynamic properties, enabling more stable and informative representations;  
(3) We propose a modality-aware Mixture-of-Experts model to facilitate feature fusion of multimodal data;  
(4) We design a cascaded diffusion framework that progressively refines human-object interaction features under dedicated supervision.  
We conduct extensive quantitative and qualitative experiments, demonstrating that MP-HOI significantly outperforms existing methods by generating motions that are not only well-aligned with textual prompts but also capable of modeling fine-grained human-object interactions.

\Acknowledgements{This work was supported by the Academic Excellence Foundation of BUAA for PhD Students, the National Natural Science Foundation of China (Project Number: 62272019), the China Postdoctoral Science Foundation (Grant Number: 2025M774236), and the Postdoctoral Fellowship Program of CPSF (Grant Number: GZC20242159).}

\bibliographystyle{plain}
\bibliography{main.bib}

\end{document}